\documentclass[letterpaper, 10 pt, conference]{ieeeconf}  

\IEEEoverridecommandlockouts                              

\usepackage{graphics} 
\usepackage{mwe}
\usepackage{subcaption}
\usepackage{graphicx}
\usepackage{multirow}

\graphicspath{ {./images/} }

\title{\LARGE \bf
Using Speech to Reduce Loss of Trust in Humanoid Social Robots*
}

\author{Amandus Krantz$^{1}$, Christian Balkenius$^{1}$, and Birger Johansson$^{1}$
\thanks{* This work was partially supported by the Wallenberg AI, Autonomous Systems and Software Program - Humanities and Society (WASP-HS) funded by the Marianne and Marcus Wallenberg Foundation and the Marcus and Amalia Wallenberg Foundation.}
\thanks{$^{1}$Lund University Cognitive Science, Department of Philosophy, Lund University, Lund, Sweden
        {\tt\small \{amandus.krantz | christian.balkenius | birger.johansson\}@lucs.lu.se}}%
}

\begin{document}

\maketitle
\thispagestyle{empty}
\pagestyle{empty}

\begin{abstract}
We present data from two online human-robot interaction experiments where 227 participants viewed videos of a humanoid robot exhibiting faulty or non-faulty behaviours while either remaining mute or speaking. The participants were asked to evaluate their perception of the robot's trustworthiness, as well as its likeability, animacy, and perceived intelligence. The results show that, while a non-faulty robot achieves the highest trust, an apparently faulty robot that can speak manages to almost completely mitigate the loss of trust that is otherwise seen with faulty behaviour. We theorize that this mitigation is correlated with the increase in perceived intelligence that is also seen when speech is present.
\end{abstract}

\section{INTRODUCTION}
As robots become integrated into our society and begin taking on a more social role by entering our homes and workplaces, understanding what it is that makes us trust those robots becomes increasingly important. Even more so, it is important to understand how trust in robots is lost and how this loss can be mitigated.

Trust and loss-of-trust mitigation in human-robot interaction (HRI) is often approached from a performance perspective \cite{hancock2011}. However, many theories of trust point out that trust also has a social component (See e.g. \cite{fiske2007, marsh1994, mcallister1995}). This social component is based more on a feeling of safety and comfort, rather than purely on rational reasoning about the system's past performance. Understanding how this more social component of trust behaves in HRI scenarios and how the behaviour of a robot may impact it is still a relatively new endeavour when compared to the more traditional performance-based trust.

Previous HRI studies have found that loss of trust or affection towards a robot that has made a mistake can be mitigated by making the robot appear more social by giving a verbal explanation of why the error happened \cite{cameron2021}. However, unlike with most neurotypical humans, the ability to speak is not a given for robots, which more often than not are completely mute. How trust in robots is affected by the ability to speak, without necessarily providing explanations for errors, is to our knowledge still an unexplored area of research. 

We designed two online independent measures experiments where participants were asked to view a video of a robot exhibiting one of two different behaviours, and afterwards evaluate their perceptions about the robot. Experiment 1 aimed to investigate how faulty and non-faulty gaze behaviours impact trust in HRI. The results of the experiment were ultimately inconclusive, showing no difference in trust between the two conditions. As faulty behaviour has been shown to affect perceptions of robots \cite{salem2013} and negatively impact trust in HRI \cite{salem2015}, we theorized the cause to be the fact that a portion of the experiment involved the robot ``speaking''. A follow-up experiment, Experiment 2, was thus performed, recreating Experiment 1 as closely as possible, but without the speech portion, this time achieving conclusive results. 

This paper thus presents data from these two HRI experiments that together may shed some light on how the ability to speak may impact the perception of a robot in HRI.

\subsection{A note on online HRI experiments}
While it may be difficult to convey some subtler elements of HRI using online studies, it is still a commonly used approach and gives access to a much larger and more diverse group of potential experiment participants compared to live-HRI experiments. At the very least, the results from such studies can be used as guidance for experiments that may be worth replicating in live HRI studies \cite{cameron2021}.

\section{METHODOLOGY}
\subsection{Participants}
The experiments were done with a total of 227 participants, 110 in Experiment 1 and 117 in Experiment 2. They were recruited from the online participant recruitment platform Prolific (prolific.co). Participants were required to be fluent in English and naïve to the purpose of the experiment (i.e., participants from Experiment 1 could not participate in Experiment 2), but otherwise no pre-screening of the participants was done. The mean age of the participants in Experiment 1 was 27 years (SD 7.73; range from 18 to 53), in Experiment 2 it was 39 years (SD 15.84; range from 18 to 75). In Experiment 1, the distribution of genders was 49.1\% identifying as male, 50\% identifying as female, and 0.9\% preferring not to say. For Experiment 2, the distribution of genders was 53.3\% identifying as male, 46.7\% identifying as female, and 0\% preferring not to say.

All participants were required to give their consent to participating in the experiment before beginning.

\subsection{Robot}
\begin{figure}
    \centering
    \includegraphics[width=0.5\linewidth]{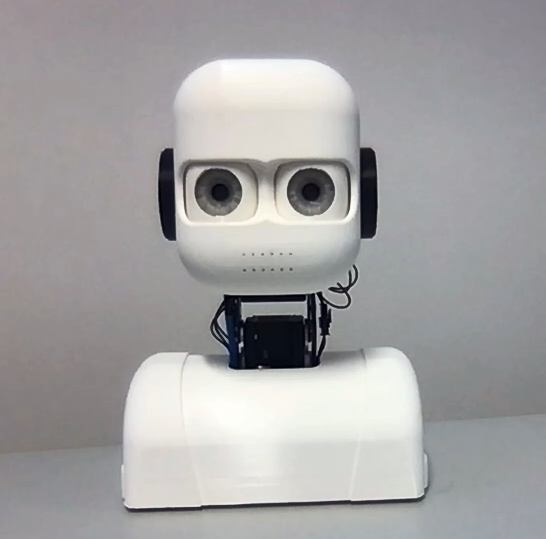}
    \caption{Epi, the humanoid robotics platform used in the experiment.}
    \label{fig:epi}
\end{figure}

The experiments were done using the humanoid robot platform Epi (See Figure \ref{fig:epi}), developed at Lund University \cite{johansson2020}. The robot's head is capable of playing pre-recorded smooth and fluid movements with 2 degrees of freedom (yaw and pitch), and has a speaker built into its ``mouth''. The eyes of the robot also have 1 degree of freedom (yaw), adjustable pupil size, and adjustable intensity of its illuminated pupils. Only movement of the head and the speaker was used for the experiments.

\subsection{Experiment set-up}
\begin{figure}
    \centering

    \begin{subfigure}{\linewidth}
        \centering

        \begin{subfigure}{0.35\linewidth}
            \includegraphics[width=\linewidth]{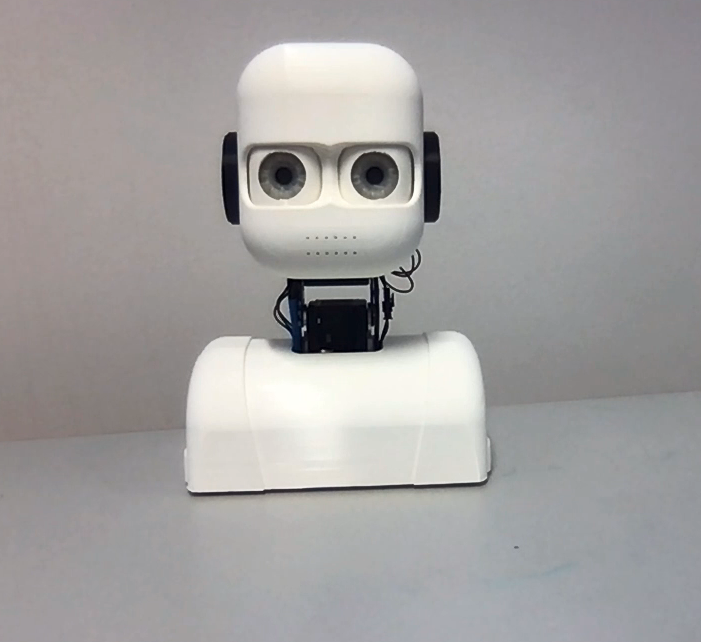}
        \end{subfigure}
        \hspace{0.5cm}
        \begin{subfigure}{0.35\linewidth}
            \includegraphics[width=\linewidth]{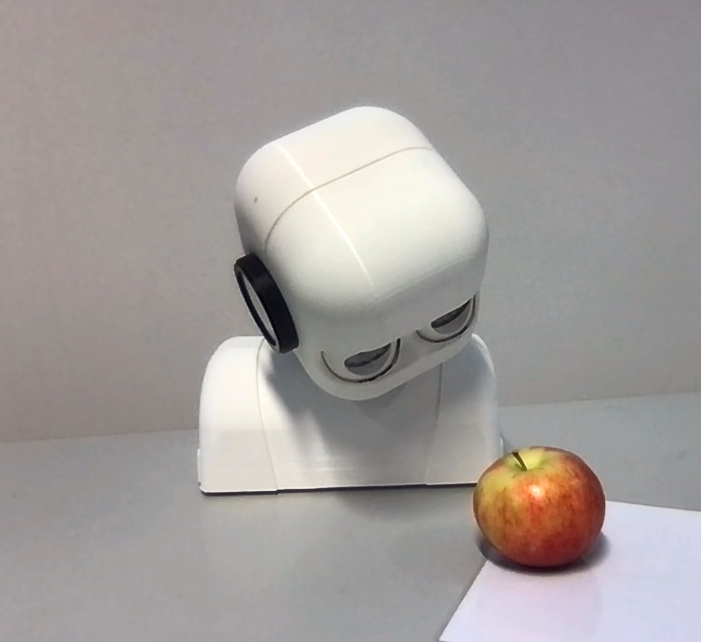}
        \end{subfigure}
        \caption{Gaze positions of non-faulty gaze behaviour.}
        \label{fig:proper}
    \end{subfigure}

    \begin{subfigure}{\linewidth}
        \centering

        \begin{subfigure}{0.35\linewidth}
            \includegraphics[width=\linewidth]{epi-camera.png}
        \end{subfigure}
        \hspace{0.5cm}
        \begin{subfigure}{0.35\linewidth}
            \includegraphics[width=\linewidth]{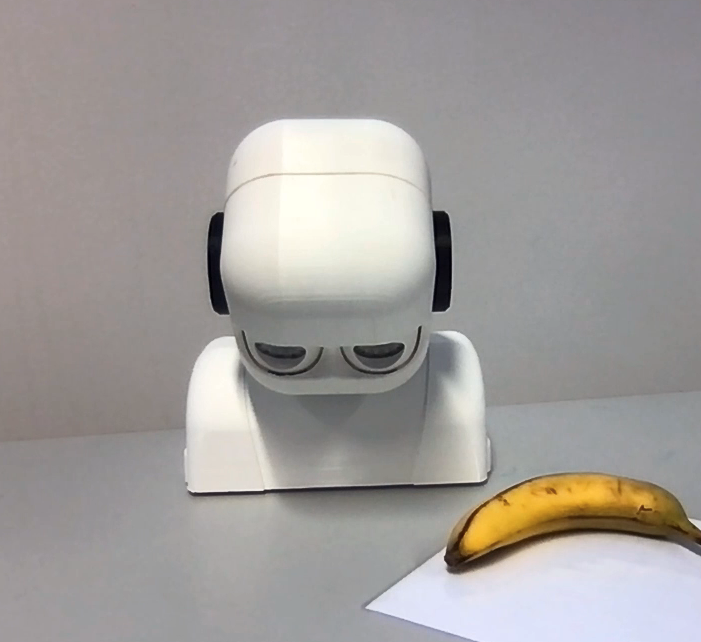}
        \end{subfigure}

        \vspace{0.5cm}

        \begin{subfigure}{0.35\linewidth}
            \includegraphics[width=\linewidth]{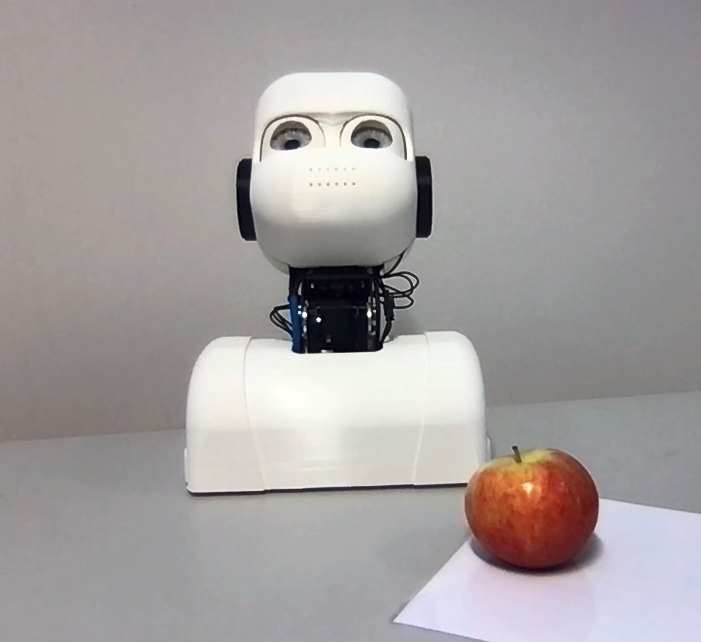}
        \end{subfigure}
        \hspace{0.5cm}
        \begin{subfigure}{0.35\linewidth}
            \includegraphics[width=\linewidth]{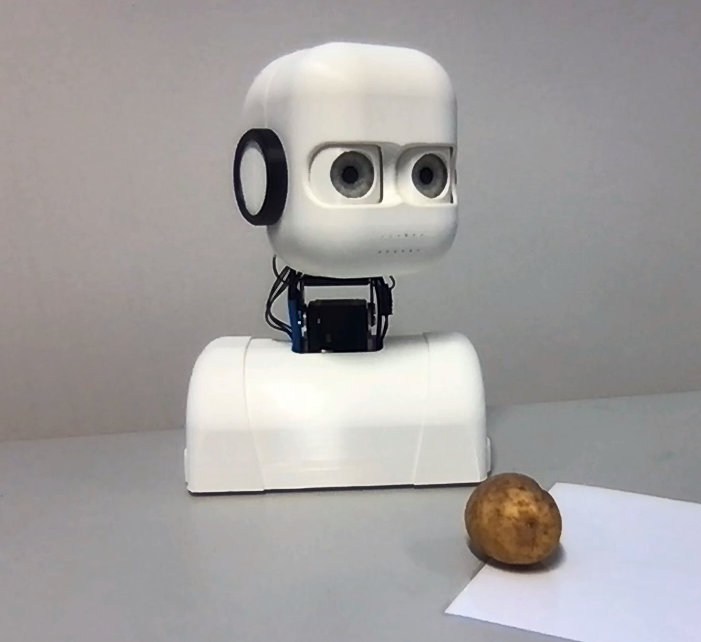}
        \end{subfigure}
        \caption{Gaze positions of faulty gaze behaviours.}
        \label{fig:faulty}
    \end{subfigure}

    \caption{Gaze positions of the different gaze behaviours in the experiments. The robot smoothly moves between looking into the camera and one of the gaze positions when presented with an object.}
    \label{fig:behaviours}
\end{figure}

Both experiments had a between-group design, where each participant was assigned to one of two conditions. Each condition had an associated video that the participants were told to base their evaluation of trust on. The videos showed the robot exhibiting either faulty or non-faulty gaze behaviours.

In the non-faulty gaze behaviour (See Figure \ref{fig:proper}), the robot starts looking into the camera. When an object is presented to the robot, the head moves until it appears to look at the object, holds the position for roughly 1 second, and moves back to its starting position, looking into the camera.

In the faulty gaze behaviour (See Figure \ref{fig:faulty}), the robot again starts looking into the camera. When the object is presented, the head moves in a random direction, rather than in the direction of the object. We chose this behaviour over having the robot remain static, as it was important that the robot appeared to have the same capabilities in all conditions. All other behaviour in the conditions with faulty gaze-behaviour is identical to the non-faulty behaviours.

In Experiment 1, once the gaze behaviour had been displayed, the robot would play a pre-recorded audio file of a computerized voice presenting a number of facts about the object that had been displayed. The speech makes no reference to whether or not the robot display a faulty or non-faulty behaviour.

Care was taken to ensure that the robot's speech never overlapped with the movement of the head. All behaviours exhibited by the robot were pre-recorded and no autonomous behaviours were implemented.

\subsection{Experiment scenario}
To avoid any observer effects, it was necessary to give the participants a scenario for which to judge the trustworthiness of the robot. As the purpose of Experiment 1 was to examine the effect of different gaze behaviours, the scenario was that the robot was being developed for a classroom setting, that its purpose was to answer children's questions, and that it was different voices we were comparing.

Experiment 2 had no speech component, so the participants were instead told that the robot was reporting which objects it was seeing to an unseen operator.

All participants were debriefed and told the real purpose of the experiment after completion.

\subsection{Measures}
Due to the dynamic nature of trust \cite{blomqvist1997, glikson2020}, we measured the amount of trust the participants felt towards the robot twice; before and after the interaction. For the pre-interaction measurement, the participants evaluated the trust based on a static image of the robot (See Figure \ref{fig:epi}). For the post-interaction measurement, they based their evaluation on one of the previously described videos. The trust relation was measured using the 14-item sub-scale of the TPS-HRI scale, developed by Schaefer et al. \cite{schaefer2016}. The scale outputs a value between 0 and 100, where 0 is complete lack of trust and 100 is complete trust.

The Godspeed scale \cite{bartneck2009}, specifically the Perceived Intelligence, Likeability, and Animacy sub-scales, were used to measure the participants' impressions of the robot after the interaction.

To control for any negative feelings the participants may have harboured towards robots before the experiment, the Negative Attitudes Towards Robots Scale (NARS) was used \cite{syrdal2009}. NARS gives an overview of both general negative feelings towards robots, and three sub-scales for negative feelings towards interaction with robots, social influence of robots, and emotions in robots.

Since robot experience has been shown to affect feelings of trust towards robots \cite{rogers2020}, we also asked the participants how often they interact with robots and autonomous systems on a 5-point scale, where 1 was daily interaction and 5 was rare or no interaction. 

\section{RESULTS}
\subsubsection{Trust}
\begin{figure}
    \includegraphics[width=\linewidth]{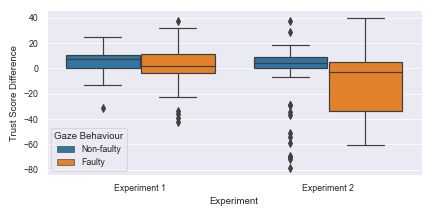}
    \caption{Box plot of differences in trust before and after interaction.}
    \label{fig:diff}
\end{figure}

In Experiment 1, no significant difference was found between the faulty and non-faulty conditions (Mann-Whitney U, $p=0.179$). However, once the speech of the robot was removed in Experiment 2, a significant difference was found (Mann-Whitney U, $p<0.05$). Looking at the box-plot of differences in trust in Figure \ref{fig:diff}, this difference seems to be due to the faulty behaviour reducing the trust, rather than the non-faulty behaviour increasing the trust. No significant difference was found between the non-faulty conditions in Experiment 1 and Experiment 2 (Mann-Whitney U, $p=0.230$).

\subsubsection{Perceived characteristics}
\begin{table}
\centering
\begin{tabular}{cllll}
\hline
\multicolumn{1}{l}{Experiment} &
  Condition &
  \multicolumn{1}{c}{Animacy} &
  \multicolumn{1}{c}{Likeability} &
  \multicolumn{1}{c}{\begin{tabular}[c]{@{}c@{}}Perceived \\ Intelligence\end{tabular}} \\ \hline
\multirow{2}{*}{Experiment 1} & Non-faulty & 2.903 & 4.011 & 4.135 \\
                              & Faulty     & 2.815 & 3.764 & 3.927 \\ \cline{2-5} 
\multirow{2}{*}{Experiment 2} & Non-faulty & 2.464 & 3.331 & 3.311 \\
                              & Faulty     & 2.244 & 2.971 & 3.036 \\ \cline{2-5} 
\end{tabular}
\caption{Mean scores from the Godspeed questionnaires for Animacy, Likability, and Perceived Intelligence.}
\label{tab:godspeed}
\end{table}

Mean scores from the Godspeed questionnaires for Animacy, Likability, and Perceived Intelligence can be found in Table \ref{tab:godspeed}. Cronbach's Alpha with a confidence interval of $0.95$ for all Godspeed questionnaires were in the $0.7 - 0.9$ interval, indicating acceptable to good internal consistency. Both conditions from Experiment 1 rank higher than Experiment 2 in all measured characteristics.

\subsubsection{Negative attitudes towards robots}
\begin{figure}
    \includegraphics[width=\linewidth]{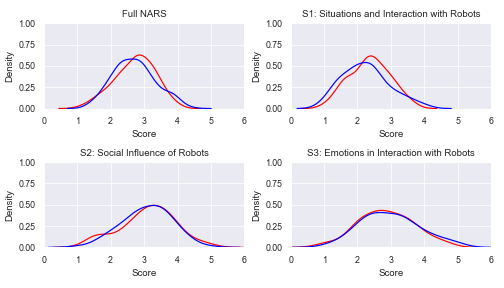}
    \caption{Kernel Density Estimate of NARS and its three sub-scales. A lower value indicates a more negative attitude. Red is Experiment 1, blue is Experiment 2.}
    \label{fig:nars}
\end{figure}

Figure \ref{fig:nars} shows the Kernel Density Estimate of NARS and its three sub-scales. The full NARS scale and the two sub-scales S2 and S3 are roughly normally distributed, indicating that the participants had overall neutral feelings towards robots before starting the experiment. The sub-scale S1 skews slightly lower, indicating that the participants had slightly negative feelings towards social situations and interactions with robots.

No significant differences can be seen in negative attitudes between the two experiments.

\subsubsection{Participants' experience with robots}
\begin{table}
\centering
\begin{tabular}{lll}
\hline
Frequency    & Experiment 1 & Experiment 2 \\ \hline
Daily        & 40\%         & 41\%         \\
Once a week  & 30\%         & 22.2\%       \\
Once a month & 13.6\%       & 14.5\%       \\
Once a year  & 8.2\%        & 11.1\%       \\
Never        & 8.2\%        & 11.1\%       \\ \hline
\end{tabular}
\caption{Proportions of how frequently the participants in either experiment interact with robots, AI, and other autonomous systems.}
\label{tab:interaction}
\end{table}

The participants in either experiment interact with robots, AI, and autonomous systems with roughly equal frequency (See Table \ref{tab:interaction}), with the majority interacting with such systems daily.

\addtolength{\textheight}{-6.5cm}

\section{DISCUSSION AND CONCLUSION}
The combined results of the two experiments show that, if the robot behaves in a non-faulty manner, unsurprisingly, trust in the robot remains largely unaffected, regardless of whether it can speak. However, once the robot is perceived as being faulty, having the ability to speak seems to reduce the resulting loss of trust, making the faulty robot appear about as trustworthy as the non-faulty ones. According to the results from the Godspeed questionnaire, the speaking robots were also perceived as being more animated, likeable, and, notably, as possessing significantly higher intelligence than the non-speaking robots. This could be an indication that, for humanoid robots, the ability to speak is perceived as a sign of high intelligence. Alternatively, the speaking robot may appear to be more sophisticated or be more capable than the non-speaking robot. Both high perceived intelligence and high capability are believed to have some correlation with a higher trust \cite{glikson2020}.

Regarding participant-centric characteristic that may affect the trust in the robot, we controlled for mean age, gender distribution, pre-existing negative attitudes towards robots, and participant experience with robots and other autonomous systems. Of these characteristics, only age differed significantly between the two experiments, with the mean age being 12 years higher in Experiment 2. While age has been shown to have an impact on attitudes towards technology, with older people having a more negative attitude \cite{degraaf2013}, the negligible difference that was seen in the distributions of the NARS scores (Figure \ref{fig:nars}) indicate that the difference in mean age between the experiments is likely not large enough to affect the results.

As ever, there are some limitations that should be kept in mind when using these results. First, as mentioned, the experiments were done online using pre-recorded videos of the robot rather than direct human-robot interaction. The large amount of available participants should safeguard against false positives, however a live-HRI study may nevertheless yield different results.

Second, the experiment scenario was different between the two experiments, with participants in Experiment 1 being told that the voice was the focus of the study. This could potentially have caused participants to ignore the gaze behaviour of the robot and focus solely on its voice, which was the same across the conditions. A follow-up study is planned to investigate this possibility.

Finally, the content of the robot's speech was not controlled for. It is conceivable that some part of the speech is signalling to some participants that the robot is highly capable or intelligent, causing the trust to increase.

In summary, this paper presents results from two experiments in HRI that together suggest that a humanoid robot with the ability to speak may not suffer the same loss of trust when displaying faulty behaviour as a robot without the ability to speak. We theorize that this effect is due to speech increasing a humanoid robot's perceived intelligence, which has been shown to correlate with trust in HRI \cite{glikson2020}. Further research along these lines may help explain existing studies in HRI (See e.g. \cite{cameron2021}) that indicate that a robot providing a verbal explanation for its errors is beneficial for user attitudes.

\bibliography{references.bib}
\bibliographystyle{IEEEtran}

\end{document}